\documentclass[sigconf]{acmart}

\usepackage{multirow}
\usepackage{diagbox}
\usepackage{graphicx}
\usepackage{balance}

\AtBeginDocument{%
  \providecommand\BibTeX{{%
    \normalfont B\kern-0.5em{\scshape i\kern-0.25em b}\kern-0.8em\TeX}}}

\copyrightyear{2022}
\acmYear{2022}
\setcopyright{acmcopyright}\acmConference[MM '22]{Proceedings of the 30th ACM
International Conference on Multimedia}{October 10--14, 2022}{Lisboa, Portugal}
\acmBooktitle{Proceedings of the 30th ACM International Conference on Multimedia
(MM '22), October 10--14, 2022, Lisboa, Portugal}
\acmPrice{15.00}
\acmDOI{10.1145/3503161.3547801}
\acmISBN{978-1-4503-9203-7/22/10}

\begin{document}

\title{Towards Adversarial Attack on Vision-Language Pre-training Models}


\author{Jiaming Zhang}
\authornote{This work is done when the author interned at Peng Cheng Lab.}
\affiliation{%
  \institution{School of Computer and Information Technology \& Beijing Key Lab of Traffic Data Analysis and Mining, Beijing Jiaotong University, China}
  \country{}
}
\email{jiamingzhang@bjtu.edu.cn}

\author{Qi Yi}
\affiliation{%
  \institution{School of Computer and Information Technology \& Beijing Key Lab of Traffic Data Analysis and Mining, Beijing Jiaotong University, China}
  \country{}
}
\email{21125273@bjtu.edu.cn}

\author{Jitao Sang}
\authornote{Corresponding authors}
\affiliation{%
  \institution{$^{1}$School of Computer and Information Technology \& Beijing Key Lab of Traffic Data Analysis and Mining, Beijing Jiaotong University, China}
  \institution{$^{2}$Peng Cheng Lab, China}
  \country{}
}
\email{jtsang@bjtu.edu.cn}

\renewcommand{\shortauthors}{Jiaming Zhang, Qi Yi, \& Jitao Sang}

\begin{abstract}

While vision-language pre-training model (VLP) has shown revolutionary improvements on various vision-language (V+L) tasks, the studies regarding its adversarial robustness remain largely unexplored. This paper studied the adversarial attack on popular VLP models and V+L tasks. First, we analyzed the performance of adversarial attacks under different settings. By examining the influence of different perturbed objects and attack targets, we concluded some key observations as guidance on both designing strong multimodal adversarial attack and constructing robust VLP models. Second, we proposed a novel multimodal attack method on the VLP models called \emph{Collaborative Multimodal Adversarial Attack} (Co-Attack), which collectively carries out the attacks on the image modality and the text modality. Experimental results demonstrated that the proposed method achieves improved attack performances on different V+L downstream tasks and VLP models. The analysis observations and novel attack method hopefully provide new understanding into the adversarial robustness of VLP models, so as to contribute their safe and reliable deployment in more real-world scenarios. Code is available at {\textbf{\href{https://github.com/adversarial-for-goodness/Co-Attack}{\url{https://github.com/adversarial-for-goodness/Co-Attack}}}}.

\end{abstract}



\maketitle

\section{Introduction}\label{sec_1}

Vision-and-Language Pre-training (VLP) has revolutionized downstream Vision-and-Language (V+L) tasks very recently~\cite{khan2021exploiting, lei2021understanding, shi2021dense}, such as image-text retrieval, visual grounding and visual entailment. This calls back attention to the research of multimodal neural networks over the last $10$ years. While extensive studies have been conducted to achieve remarkable progress, only a few of them have investigated the adversarial robustness problem, which typically selects one single modality for perturbation to attack the multimodal task with standard adversarial attack method~\cite{shah2019cycle, xu2018fooling, yang2021defending}. Regarding multimodal pre-training especially VLP models, so far as we know, there has been no study to systematically analyze the adversarial robustness performance and design a dedicated adversarial attack solution.


\begin{figure}[t]
  \centering
  \includegraphics[width=\linewidth]{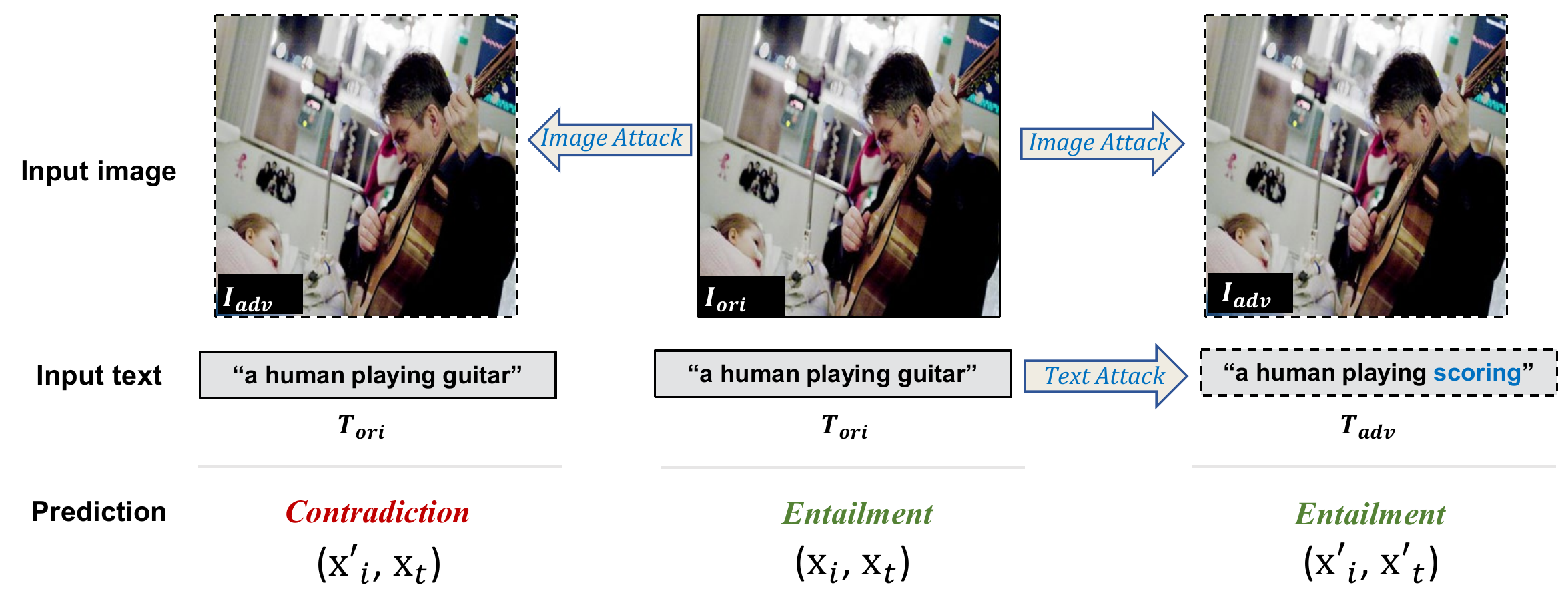}
  \caption{An example of adversarial attack on the VLP model in visual entailment task.}\label{fig_1}
\end{figure}

\begin{table*}[t]
\centering
\caption{Attack settings along different attack targets and perturbed objects.}\label{tab_1}
\begin{tabular}{|lc|c|c|c|c|}
\hline
 \multicolumn{2}{|c|}{\multirow{2}*{\diagbox{Perturbed Object}{Attack Target}}}& \multicolumn{2}{c|}{Unimodal Embedding}  & \multicolumn{2}{c|}{Multimodal Embedding} \\
 \cline{3-6}
  & & {\verb|[CLS]|} of embedding  & full embedding & {\verb|[CLS]|} of embedding & full embedding \\
 \hline
 \multirow{2}*{Single-modal input} & Image  & \texttt{Image@Uni$\rm_{CLS}$} & \texttt{Image@Uni$\rm_{full}$} & \texttt{Image@Multi$\rm_{CLS}$} & \texttt{Image@Multi$\rm_{full}$} \\
 \cline{2-6}
 & Text & \texttt{Text@Uni$\rm_{CLS}$} & \texttt{Text@Uni$\rm_{full}$} & \texttt{Text@Multi$\rm_{CLS}$} & \texttt{Text@Multi$\rm_{full}$} \\
 \hline
 Bi-modal inputs & Image\&Text & \texttt{Bi@Uni$\rm_{CLS}$} & \texttt{Bi@Uni$\rm_{full}$} & \texttt{Bi@Multi$\rm_{CLS}$} & \texttt{Bi@Multi$\rm_{full}$} \\
 \hline
\end{tabular}
\end{table*}

As still in its very early stage, many problems remain unexplored regarding the adversarial attack on VLP models. Among these, two critical issues are examined in this paper: (1) Standard adversarial attack is designed for classification tasks involved with only one modality. VLP models involve multiple modalities and typically many non-classification tasks like image-text cross-modal retrieval, making it impractical to directly employ the standard adversarial attack methods. A natural solution is to conduct adversarial attack on the embedding representation instead of the downstream task labels. However, due to the complex structure of VLP embedding representation, the problem turns to understanding how different attack settings affect the attack performance. (2) To attack the embedding representation of VLP model, the adversarial perturbations for different modalities should be considered collaboratively instead of independently. Figure~\ref{fig_1} illustrates an example regarding adversarial attack on ALBEF~\cite{li2021align} for the visual entailment task. It is shown that only perturbing image can successfully change the prediction from ``entailment'' to ``contradiction''. By independently perturbing both image and text without considering their interaction, however, the attack fails as the two single-modal attacks may work in conflict with each other and lead to a countervailing $1+1<1$ effect. This study is devoted towards adversarial attack on VLP models, addressing these two issues by analyzing attack performance in different settings and developing collaborative multimodal adversarial attack solution.


For the first issue, we analyzed the adversarial attack on VLP models in different settings along two dimensions of attack target and perturbed object. Two typical VLP architectures, fused VLP models (e.g., ALBEF~\cite{li2021align}, TCL~\cite{yang2022vision}) and aligned VLP models (e.g., CLIP~\cite{radford2021learning}) are examined with $3$ downstream V+L tasks of image-text retrieval, visual entailment, and visual grounding. The extensive analyses derive some key observations regarding the different attacking performances among VLP architectures and V+L tasks, as well as the influence of attack settings on VLP robustness.


For the second issue, we propose a novel multimodal adversarial attack method on the VLP models called \emph{Collaborative Multimodal Adversarial Attack} (Co-Attack), which collectively carries out the attack on the image modality and the text modality. Co-Attack is applicable to both fused VLP and aligned VLP models. The premise is to encourage the perturbed multimodal embedding away from the original multimodal embedding (for fused VLP models), or the perturbed image-modal embedding away from the perturbed text-modal embedding (for aligned VLP models). Experimental results demonstrate that the proposed method achieves improved attack performance on different V+L downstream tasks.

The contributions of the paper can be summarized as follows. (1) We analyzed the performance of adversarial attacks on two typical architectures of VLP models and three downstream V+L tasks. The observations regarding attacking settings along attack targets and perturbed objects shed light on understanding the adversarial robustness of VLP models. (2) We developed a novel multimodal adversarial attack method on the VLP models. By considering the consistency between attacks of different modalities, it collaboratively combines multimodal perturbations towards a stronger adversarial attack.

\section{Preliminaries and Related Work}

\subsection{VLP models and Downstream V+L Tasks}

\subsubsection{VLP models}
$\\$
Most early work on vision-language representation learning required pre-trained object detectors (e.g., Faster R-CNN~\cite{ren2015faster}) and high resolution images~\cite{tan2019lxmert, li2020oscar}. Rather than relying on visual features derived by computation-costly object detectors, recent methods use an end-to-end image encoder (e.g., ViT~\cite{dosovitskiy2020image}) to speed up inference~\cite{radford2021learning, li2021align, yang2022vision}. In this work, we consider CLIP~\cite{radford2021learning}, ALBEF~\cite{li2021align} and TCL~\cite{yang2022vision} for evaluation, which belong to the latter class of models. Among them, ALBEF and TCL model the interactions between image and text modality, containing both the unimodal encoder $E_i(\cdot)$, $E_t(\cdot)$, and a multimodal encoder $E_m(\cdot, \cdot)$ (illustrated in Figure~\ref{fig_2}(a)). An input image $\mathbf{x}_i$ is encoded into the image embedding $e_i$ by the image encoder $E_i(\cdot)$, i.e., $e_i=E_i(\mathbf{x}_i)$. An input text $\mathbf{x}_t$ is encoded into the text embedding $e_t$ by the text encoder $E_t(\cdot)$, i.e., $e_t=E_t(\mathbf{x}_t)$. The image embedding $e_i$ and text embedding $e_t$ are then fed into the multimodal encoder $E_m(\cdot, \cdot)$ to output a unified multimodal embedding $e_m$, i.e., $e_m=E_i(e_i, e_t)$. We denote this type of VLP models with multimodal encoder and unified multimodal embedding as fused VLP model. In contrast, CLIP focuses on learning unimodal image encoder and text encoder without considering the multimodal encoder. We denote this type of VLP models with only separate unimodal embeddings as the aligned VLP model (illustrated in Figure~\ref{fig_2}(b)).

\subsubsection{Downstream V+L Tasks}
$\\$
\textbf{Image-Text Retrieval} contains two sub-tasks: image-to-text retrieval (TR) and text-to-image retrieval (IR). For ALBEF and TCL, both for TR and IR, the feature similarity score is calculated at first between $e_i$ and $e_t$ for all image-text pairs to retrieve the Top-N candidates, and then the image-text matching score calculated by $e_m$ is used for ranking. TR and IR tasks on CLIP are carried out more directly. The ranking results are only based on the similarity between $e_i$ and $e_t$.

\textbf{Visual Entailment (VE)} is a visual reasoning task to predict whether the relationship between an image and a text is entailment, neutral, or contradiction. Both ALBEF and TCL consider VE as a three-way classification problem, and predict the class probabilities using a fully layer on the multimodal encoder’s representation of the {\verb|[CLS]|} token~\cite{li2021align}.

\textbf{Visual Grounding (VG)} localizes the region in an input image based on the description of the corresponding input text. ALBEF extends Grad-CAM~\cite{selvaraju2017grad}, and uses the derived attention map to rank the detected proposals~\cite{yu2018mattnet}.

\subsection{Adversarial Attack}

\subsubsection{In Computer Vision and Natural Language Processing}
$\\$
Adversarial attack is first proposed in computer vision, which demonstrates the vulnerability of deep learning models~\cite{szegedy2013intriguing}. The most common adversarial attacks in computer vision are gradient-based methods such as FGSM~\cite{goodfellow2014explaining}, PGD~\cite{madry2017towards}, MIM~\cite{dong2018boosting} and SI~\cite{lin2020nesterov}. To obtain the image adversarial perturbations $\delta_i$ corresponding to input image $\mathbf{x}_i$, we consider $\ell_\infty$-norm constrained perturbation in this work, where $\delta_i$ satisfies $\Vert \delta_i \Vert_\infty \leq \epsilon_i$ with $\epsilon_i$ being the maximum perturbation magnitude.

In computer vision, adversarial attack can be easily achieved by gradient-based methods in the continuous space. Unlike this, in natural language processing, adversarial attack in the discrete space is still a challenge. Current successful adversarial attacks in natural language processing usually modify or replace some tokens of input text $\mathbf{x}_t$ to maximize the risk of outputting wrong embedding~\cite{ren2019generating, li2020bert}. 

\subsubsection{In Multimodal Neural Network}
$\\$
Regarding VLP models, so far as we know, there has been no related work to systematically analyze and design adversarial attack. Therefore, we discuss and employ these available adversarial attacks on multimodal neural networks involving V+L tasks as the baseline methods for comparison. \citet{xu2018fooling} investigated attacking the visual question answering model by perturbing the image modality. \citet{agrawal2018don} and \citet{shah2019cycle} attempted to attack the vision-and-language model by perturbing the text modality. \citet{yang2021defending} studied the robustness of various multimodal models and proposed a defense method. We summarize three main differences between these multimodal attack studies and this work. First, the models in the above studies are based on the convolutional neural networks (CNNs) and recurrent neural networks (RNNs). The methods and observations are not readily applicable to the transformer-based VLP models. Second, the above studies mainly addressed single V+L classification tasks, and the attacks used are not generalized to other non-classification tasks. Third, the adversarial attack methods they employed are essentially standard unimodal attack, while multimodal attacks specially for multimodal models remain unexplored.

\section{Analyzing Adversarial Attack in VLP Model}\label{sec_3}


\subsection{Analysis Settings}\label{sec_31}

\subsubsection{Examined VLP Models, Downstream Tasks and Datasets}

\paragraph{\textbf{Examined VLP Models}}
We evaluated the analyses and experiments on two types of models: the fused VLP model and the aligned VLP model. The fused VLP model contains an image encoder, a text encoder, and a multimodal encoder, and in this work we consider ALBEF~\cite{li2021align} and TCL~\cite{yang2022vision}. The image encoder is implemented by a 12-layer visual transformer ViT-B/16~\cite{dosovitskiy2020image}. Both the text encoder and the multimodal encoder are implemented by a 6-layer transformer. The aligned VLP model contains only image encoder and text encoder, and in this work we consider CLIP. There are different choices of image encoder for CLIP. We consider ViT-B/16 and ResNet-101~\cite{he2016deep}, denoted as CLIP$_{\rm ViT}$ and CLIP$_{\rm CNN}$, respectively.

\paragraph{\textbf{Downstream Tasks and Datasets}}
In this work, MSCOCO~\cite{lin2014microsoft} and Flickr30K~\cite{plummer2015flickr30k} are used to evaluate TR and IR tasks, and RefCOCO+~\cite{yu2016modeling} is used to evaluate VG task, and SNLI-VE~\cite{xie2019visual} is used to evaluate VE task. Note that CLIP can only tackle IR and TR tasks, TCL can handle VE, IR and TR tasks, and ALBEF can deal with all the above downstream tasks. For the VE task, since we are concerned with the performance of adversarial attack, we select only positive image-text pairs (with label of \emph{entailment}) and discard the negative pairs (with label of \emph{neutral} and \emph{contradictory}) from the SNLI-VE testing dataset.

\paragraph{\textbf{Hyper-parameters}}
For the adversarial attack on image modality, we use the PGD attack~\cite{madry2017towards}. The maximum perturbation $\epsilon_i$ is set to $2/255$. The step size is set to $1.25$. The number of iterations is set to $10$. For the adversarial attack on text modality, we use BERT-Attack~\cite{li2020bert}. The maximum perturbation $\epsilon_t$ is set to $1$ token. The length of the selected word list is set to $10$.

\begin{figure}[t]
  \centering
  \includegraphics[width=0.7\linewidth]{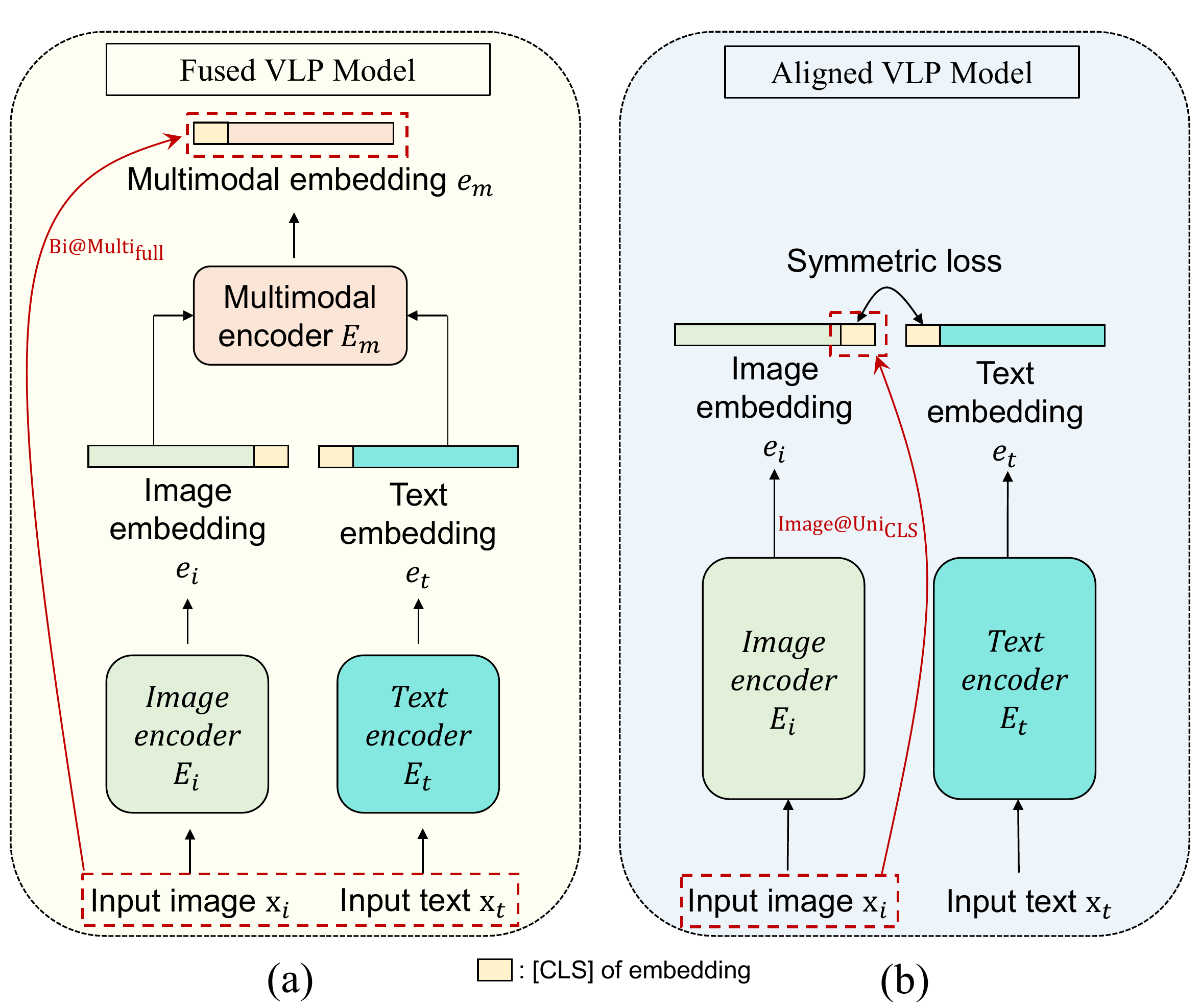}
  \caption{Illustration of VLP model architectures and the attack settings. (a) The fused VLP model consists of an image encoder, and a text encoder and a multimodal encoder. (b) The aligned VLP model has no multimodal encoder and no unified multimodal embedding. The two red arrows point from the perturbed object to the attack target.}\label{fig_2}
\end{figure}

\subsubsection{Attack Targets and Perturbed Objects}

Table~\ref{tab_1} lists the attack targets and perturbation objects that are considered in this work. (1) For the perturbation objects of the VLP model, we consider three options: image-modal input, text-modal input and bi-modal (image\&text) inputs. (2) For the attack targets, they can be coarsely classified into two types: multimodal embedding outputted from the multimodal encoder and unimodal embedding outputted from the image or text encoders. Within each type of the attacked embedding, it is further divided into the full embedding and {\verb|[CLS]|} of embedding. The red arrows in Figure~\ref{fig_2} illustrate two example attack settings: \texttt{Bi@Multi$\rm_{full}$}, perturbing the input image and input text simultaneously to attack the full multimodal embedding; \texttt{Image@Uni$\rm_{CLS}$}, perturbing the input image to attack the {\verb|[CLS]|} of unimodal embedding.

We briefly discuss the motivation to set {\verb|[CLS]|} of embedding as one attack target. {\verb|[CLS]|} of embedding plays an important role in the pre-training models, e.g., the {\verb|[CLS]|} of embedding in the VLP model is directly employed to inference in various downstream tasks. Therefore, it is worth examining the effectiveness of attacking {\verb|[CLS]|} of embedding in VLP models. Note that the distinction between {\verb|[CLS]|} of embedding and full embedding does not fit for CLIP, since CLIP can replace the ViT of the image encoder with a CNN. Therefore, we only discuss {\verb|[CLS]|} of embedding for CLIP$_{\rm ViT}$ and regard the embedding outputted by CNN as {\verb|[CLS]|} of embedding in the rest of the paper.

\begin{table*}[t]
\centering
\caption{Image-text retrieval results of ALBEF on Flickr30K and MSCOCO dataset. The reported value is attack success rate where a larger value indicates a stronger attack.}\label{tab_2}
\setlength{\tabcolsep}{2.7mm}{
\begin{tabular}{c|ccc|ccc|ccc|ccc} 
\toprule
 \multirow{3}*{Attack} & \multicolumn{6}{c|}{Flickr30K (1K test set)} & \multicolumn{6}{c}{MSCOCO (5K test set)} \\
\cline{2-13}
  & \multicolumn{3}{c|}{TR} & \multicolumn{3}{c|}{IR} & \multicolumn{3}{c|}{TR} & \multicolumn{3}{c}{IR}\\
\cline{2-13}
  & R@1 & R@5 & R@10 & R@1 & R@5 & R@10 & R@1 & R@5 & R@10 & R@1 & R@5 & R@10 \\
\hline
 {\texttt{Text@Uni$\rm_{CLS}$}} & 10.10 & 1.70 & 1.10 & 21.42 & 13.84 & 10.78 & 24.50 & 15.94 & 10.46 & 26.19 & 24.95 & 21.43 \\
 {\texttt{Image@Uni$\rm_{CLS}$}} &  48.70 & 36.30 & 29.40 & 48.88 & 42.78 & 37.08 & 48.56 & 48.25 & 43.18 & 40.32 & 47.11 & 45.56 \\
 {\texttt{Bi@Uni$\rm_{CLS}$}} & 64.80 & 50.60 & 43.20 & 62.40 & 57.80 & 52.90 & 63.40 & 66.18 & 61.54 & 51.15 & 63.11 & 63.02 \\
\hline
 {\texttt{Text@Multi$\rm_{CLS}$}} & 13.70 & 2.70 & 1.30 & 24.02 & 15.10 & 11.46 & 26.82 & 16.92 & 11.50 & 27.74 & 25.78 & 21.75 \\
 {\texttt{Image@Multi$\rm_{CLS}$}} & 20.50 & 13.90 & 11.20 & 27.36 & 25.16 & 22.32 & 23.50 & 27.68 & 25.22 & 24.43 & 33.68 & 34.34 \\
 {\texttt{Bi@Multi$\rm_{CLS}$}} & 35.90 & 24.30 & 19.70 & 44.54 & 41.34 & 36.86 & 41.86 & 41.62 & 37.66 & 38.10 & 48.10 & 48.24 \\
\hline
 {\texttt{Text@Uni$\rm_{full}$}} & 7.30 & 1.10 & 0.60 & 16.02 & 10.72 & 8.64 & 15.86 & 9.40 & 6.24 & 19.70 & 18.85 & 16.49 \\
{\texttt{Image@Uni$\rm_{full}$}}& 59.20 & 48.60 & 42.70 & 56.54 & 54.96 & 50.54 & 58.40 & 61.88 & 59.18 & 46.57 & 58.41 & 58.98 \\
{\texttt{Bi@Uni$\rm_{full}$}} & 66.40 & 55.00 & 49.60 & 63.76 & 62.68 & 58.18 & 65.02 & 69.56 & 66.34 & 51.79 & 66.19 & 67.35 \\
\hline
{\texttt{Text@Multi$\rm_{full}$}}& 7.70 & 1.40 & 1.00 & 16.78 & 11.40 & 8.98 & 15.00 & 8.72 & 5.74 & 19.11 & 18.39 & 15.92 \\
{\texttt{Image@Multi$\rm_{full}$}}& 60.00 & 50.20 & 42.90 & 58.74 & 56.90 & 52.02 & 54.30 & 58.58 & 55.32 & 46.77 & 60.26 & 60.94 \\
{\texttt{Bi@Multi$\rm_{full}$}}& 65.70 & 55.40 & 48.10 & 62.92 & 62.80 & 58.28 & 60.54 & 65.78 & 62.72 & 51.08 & 65.86 & 67.09 \\
\bottomrule
\end{tabular}}
\end{table*}

\subsection{Attack Implementations}

\paragraph{\textbf{Attacking unimodal embedding}}

For perturbing the image-modal input, most of the typical unimodal adversarial attacks on classification are gradient-based methods. Among them, we selected FGSM\cite{goodfellow2014explaining}, which requires only one gradient calculation, to represent this family of methods:
\begin{equation}\label{eq1}
  \delta_i = \epsilon_i \cdot {\rm{sign}}(\nabla_{\mathbf{x}_i} \mathcal{L}_c (C(\mathbf{x}_i), \hat{y})),
\end{equation}
where $C$ denotes the classification model, $\mathcal{L}_c$ is cross-entropy loss, and $\hat{y}$ is ground-truth label corresponding to the original image $\mathbf{x}_i$. The cross-entropy loss requires logit-wise representation, but many V+L downstream tasks are non-classification tasks, such as image-text retrieval. So we used the method proposed by \citet{zhang2019theoretically} that maximizes KL (Kullback-Leibler) divergence loss $\mathcal{L}$ of embedding-wise representation to conduct adversarial attack:
\begin{equation}\label{eq3}
  \delta_i = \epsilon_i \cdot {\rm{sign}}(\nabla_{\mathbf{x}^{'}_i} \mathcal{L} (E_i(\mathbf{x}^{'}_i), E_i(\mathbf{x}_i)).
\end{equation}
For perturbing the text-modal input, $T(\cdot)$ represents the modification or replacement of tokens in the input text $\mathbf{x}_t$, i.e., $\mathbf{x}^{'}_t = T(\mathbf{x}_t)$. Then the text adversarial perturbation $\delta_t$ can be represented as follows:
\begin{equation}\label{eq2}
  \delta_t = \mathop{\arg\max}\limits_{\mathbf{x}^{'}_t} (\Vert E_t(\mathbf{x}^{'}_t) - E_t(\mathbf{x}_t) \Vert) - \mathbf{x}_t,
\end{equation}
where the maximum perturbation $\epsilon_t$ is constrained to the token level, i.e., how many tokens are modified/replaced with semantic-consistent. In this work, we use BERT-Attack~\cite{li2020bert} to conduct adversarial attack on text modality.


\paragraph{\textbf{Attacking multimodal embedding}}
For the attack on the input text, we substitute the text embedding $E_t(\cdot)$ in Equation~\eqref{eq2} with the multimodal embedding $E_m(\cdot, \cdot)$:
\begin{equation}\label{eq4}
  \delta_t = \mathop{\arg\max}\limits_{\mathbf{x}^{'}_t} (\Vert E_m(E(\mathbf{x}_i), E_t(\mathbf{x}^{'}_t)) - E_m(E(\mathbf{x}_i), E_t(\mathbf{x}_t)) \Vert) - \mathbf{x}_t.
\end{equation}
Similarly, for the attack on the input image, we substitute the image embedding $E_i(\cdot)$ in Equation~\eqref{eq3} with the multimodal embedding $E_m(\cdot, \cdot)$:
\begin{equation}\label{eq5}
  \delta_i = \epsilon_i \cdot {\rm{sign}}(\nabla_{\mathbf{x}^{'}_i} \mathcal{L}(E_m(E_i(\mathbf{x}^{'}_i), E_t(\mathbf{x}_t)) , E_m(E_i(\mathbf{x}_i), E_t(\mathbf{x}_t))).
\end{equation}

\subsection{Observations}

\begin{table*}[t]
  \centering
  \caption{Image-text retrieval results of CLIP$_{\rm ViT}$ on Flickr30K and MSCOCO dataset. The reported value is attack success rate.}\label{tab_3}
  \setlength{\tabcolsep}{2.7mm}{
  \begin{tabular}{c|ccc|ccc|ccc|ccc}
  
  \toprule
  \multirow{3}*{Attack} & \multicolumn{6}{c|}{Flickr30K (1K test set)} & \multicolumn{6}{c}{MSCOCO (5K test set)} \\
\cline{2-13}
  & \multicolumn{3}{c|}{TR} & \multicolumn{3}{c|}{IR} & \multicolumn{3}{c|}{TR} & \multicolumn{3}{c}{IR}\\
 \cline{2-13}
  & R@1 & R@5 & R@10 & R@1 & R@5 & R@10 & R@1 & R@5 & R@10 & R@1 & R@5 & R@10 \\

 \hline
  \texttt{Text@Uni$\rm_{CLS}$}& 19.60 & 10.50 & 6.30 & 20.90 & 19.00 & 14.90 & 24.92 & 25.26 & 22.08 & 16.61 & 23.68 & 24.34 \\
  \texttt{Image@Uni$\rm_{CLS}$}& 56.20 & 49.90 & 42.20 & 43.68 & 47.00 & 42.58 & 41.34 & 52.68 & 52.74 & 26.28 & 40.79 & 44.47 \\
  \texttt{Bi@Uni$\rm_{CLS}$}& 64.00 & 60.90 & 52.90 & 51.08 & 60.34 & 58.54 & 46.34 & 62.10 & 64.34 & 29.75 & 48.78 & 54.88  \\
 \bottomrule
\end{tabular}}
\end{table*}

\begin{table*}[t]
  \centering
  \caption{Image-text retrieval results of CLIP$_{\rm CNN}$ on Flickr30K and MSCOCO dataset. The reported value is attack success rate.}\label{tab_4}
  \setlength{\tabcolsep}{2.7mm}{
  \begin{tabular}{c|ccc|ccc|ccc|ccc}
  \toprule
  \multirow{3}*{Attack} & \multicolumn{6}{c|}{Flickr30K (1K test set)} & \multicolumn{6}{c}{MSCOCO (5K test set)} \\
\cline{2-13}
  & \multicolumn{3}{c|}{TR} & \multicolumn{3}{c|}{IR} & \multicolumn{3}{c|}{TR} & \multicolumn{3}{c}{IR}\\
 \cline{2-13}
  & R@1 & R@5 & R@10 & R@1 & R@5 & R@10 & R@1 & R@5 & R@10 & R@1 & R@5 & R@10 \\

 \hline
  \texttt{Text@Uni$\rm_{CLS}$}& 18.50 & 10.90 & 6.30 & 20.40 & 20.46 & 17.14 & 22.40 & 24.00 & 22.16 & 15.77 & 24.08 & 25.25\\
  \texttt{Image@Uni$\rm_{CLS}$}& 63.30 & 65.60 & 59.70 & 51.58 & 64.24 & 63.70 & 43.12 & 58.62 & 61.62 & 27.56 & 46.67 & 52.87\\
  \texttt{Bi@Uni$\rm_{CLS}$}& 70.50 & 73.50 & 69.10 & 54.10 & 70.96 & 71.20 & 46.68 & 65.62 & 70.82 & 28.97 & 50.74 & 58.95\\
 \bottomrule
  \end{tabular}}
\end{table*}

\begin{table*}[t]
  \centering
  \caption{Visual entailment results of ALBEF and TCL on SNLI-VE dataset. The reported value is attack success rate.}\label{tab_5}
  \setlength{\tabcolsep}{0.2mm}{
  \begin{tabular}{cccc|ccc|ccc|ccc}
  \toprule
    Model  & \rotatebox{50}{\texttt{Text@Uni$\rm_{CLS}$}} & \rotatebox{50}{\texttt{Image@Uni$\rm_{CLS}$}} & \rotatebox{50}{\texttt{Bi@Uni$\rm_{CLS}$}} & \rotatebox{50}{\texttt{Text@Multi$\rm_{CLS}$}} & \rotatebox{50}{\texttt{Image@Multi$\rm_{CLS}$}} & \rotatebox{50}{\texttt{Bi@Multi$\rm_{CLS}$}} & \rotatebox{50}{\texttt{Text@Uni$\rm_{full}$}} & \rotatebox{50}{\texttt{Image@Uni$\rm_{full}$}} & \rotatebox{50}{\texttt{Bi@Uni$\rm_{full}$}} & \rotatebox{50}{\texttt{Text@Multi$\rm_{full}$}} & \rotatebox{50}{\texttt{Image@Multi$\rm_{full}$}} & \rotatebox{50}{\texttt{Bi@Multi$\rm_{full}$}} \\
    \hline
    ALBEF & 26.65& 25.52 & 38.99 & 34.36 & 39.51 & 50.90 & 20.74 & 31.78 & 39.53 & 33.17 & 42.28 & 52.72 \\
    TCL & 28.34 & 29.78 & 42.39 & 37.00 & 32.33 & 48.87 & 27.20 & 18.98 & 36.60 & 37.23 & 32.91 & 49.92 \\
\bottomrule
\end{tabular}}
\end{table*}

\paragraph{\textbf{Observations on Image-Text Retrieval}}

The results of ALBEF on the TR and IR tasks are shown in Table~\ref{tab_2}. In Table~\ref{tab_2}, we have the following main findings: (1) Perturbing bi-modal inputs (\texttt{Bi@}) is consistently stronger than perturbing any single-modal input (both \texttt{Text@} and \texttt{Image@}). This suggests that stronger adversarial attacks are expected if multiple modal inputs are allowed to be perturbed. (2) For perturbing image-modal input, attacking full embedding outperforms attacking {\verb|[CLS]|} of embedding. But for perturbing text-modal input, we observed the opposite result. This suggests that for text encoders, the {\verb|[CLS]|} of embedding represents the sentence-level significance, and attacking it has a greater impact than attacking the full embedding. However, for image encoders, attacking the full embedding has a more significant impact, which is consistent with the observation in \cite{mao2021towards}. They replaced the {\verb|[CLS]|} of embedding with a ``global average pooling'', and found that removing the {\verb|[CLS]|} of embedding contributes a negligible influence on accuracy and robustness. (3) The attack performance of \texttt{Bi@Multi$\rm_{full}$} is much better than \texttt{Bi@Multi$\rm_{CLS}$}. This indicates that for perturbing bi-modal inputs to attack multimodal embedding, the influence of {\verb|[CLS]|} of embedding is limited and even weakens the adversarial attack. (4) \texttt{Bi@Multi} and \texttt{Bi@Uni} achieve the similar performance. This suggests that for tasks like TR and IR that require intermediate outputs (unimodal embedding) involved for inference, the impact of attacking unimodal embedding and attacking full embedding is comparable.

The results of attacking CLIP$_{\rm ViT}$ and CLIP$_{\rm CNN}$ are shown in Table~\ref{tab_3} and Table~\ref{tab_4}, respectively. Main observations include: (1) It is found that despite the large difference in the architecture, consistent with the observation in ALBEF, perturbing bi-modal inputs is stronger than perturbing any single-modal input. (2) Both for perturbing image-modal input and perturbing bi-modal inputs, the attack success rate on CLIP$_{\rm CNN}$ (Table~\ref{tab_4}) is higher than CLIP$_{\rm ViT}$ (Table~\ref{tab_3}). But it is hard to tell the difference between CLIP$_{\rm CNN}$ and CLIP$_{\rm ViT}$ for perturbing text-modal input. This suggests that ViT is a more robust image encoder than ResNet-101 (CNN) on defending the image-modal attack, which is also consistent with some existing observations~\cite{shao2021adversarial}. (3) Note that the image encoder for ALBEF (Table~\ref{tab_2}) and CLIP$_{\rm ViT}$ (Table~\ref{tab_3}) are the same (ViT-B/16). By comparing the differences in the attack success rates between them, we can observe that there is no distinct superiority or inferiority between attacking the unimodal embedding of the ALBEF and CLIP$_{\rm ViT}$. This demonstrates that regarding different VLP models, different pre-training objectives of ALBEF (fused VLP model) and CLIP$_{\rm ViT}$ (aligned VLP model) have no significant impact on the adversarial robustness.

\paragraph{\textbf{Observations on Visual Entailment}}

The results of ALBEF and TCL on the VE task are shown in Table~\ref{tab_5}. We have the following main findings: (1) Consistent with the results on the image-text retrieval task, perturbing bi-modal inputs is stronger than perturbing any single-modal input, and attacking full embedding outperforms attacking {\verb|[CLS]|} of embedding for perturbing image-modal input. (2) \texttt{Bi@Uni$\rm_{CLS}$} and \texttt{Bi@Uni$\rm_{full}$} achieved the neck-and-neck performances, as did \texttt{Bi@Multi$\rm_{CLS}$} and \texttt{Bi@Multi$\rm_{full}$}. This indicates that for VE task, {\verb|[CLS]|} barely interferes with the attack performance of perturbing bi-modal inputs. (3) \texttt{Bi@Multi$\rm_{CLS}$} outperforms \texttt{Bi@Uni$\rm_{CLS}$}, and \texttt{Bi@Multi$\rm_{full}$} outperforms \texttt{Bi@Uni$\rm_{full}$}. This suggests that for perturbing bi-modal inputs, attacking multimodal embedding is substantially stronger than attacking unimodal embedding. Therefore, the impact of attacking unimodal embeddings is relatively weak for tasks like VE that do not require intermediate outputs (unimodal embedding) involved for inference.


\begin{table*}[t]
\centering
\caption{Visual grounding results of ALBEF on RefCOCO+ dataset. The reported value is attack success rate.}\label{tab_6}
\setlength{\tabcolsep}{0.2mm}{
\begin{tabular}{cccc|ccc|ccc|ccc}
\toprule
    Subset& \rotatebox{50}{\texttt{Text@Uni$\rm_{CLS}$}} & \rotatebox{50}{\texttt{Image@Uni$\rm_{CLS}$}} & \rotatebox{50}{\texttt{Bi@Uni$\rm_{CLS}$}} & \rotatebox{50}{\texttt{Text@Multi$\rm_{CLS}$}} & \rotatebox{50}{\texttt{Image@Multi$\rm_{CLS}$}} & \rotatebox{50}{\texttt{Bi@Multi$\rm_{CLS}$}} & \rotatebox{50}{\texttt{Text@Uni$\rm_{full}$}} & \rotatebox{50}{\texttt{Image@Uni$\rm_{full}$}} & \rotatebox{50}{\texttt{Bi@Uni$\rm_{full}$}} & \rotatebox{50}{\texttt{Text@Multi$\rm_{full}$}} & \rotatebox{50}{\texttt{Image@Multi$\rm_{full}$}} & \rotatebox{50}{\texttt{Bi@Multi$\rm_{full}$}} \\
    \hline
    Val & 10.45 & 5.313 & 14.85 & 9.27 & 12.81 & 16.77 & 8.44 & 12.03 & 16.50 & 9.82 & 15.97 & 19.21 \\
    
    TestA & 12.08 & 6.00 & 17.35 & 11.84 & 14.30 & 19.96 & 10.18 & 12.55 & 18.87 & 10.79 & 16.59 & 22.33 \\
    
    TestB & 7.18 & 3.50 & 10.41 & 6.97 & 10.04 & 10.96 & 5.17 & 10.04 & 11.70 & 6.81 & 11.94 & 13.99 \\
\bottomrule
\end{tabular}}
\end{table*}

\paragraph{\textbf{Observations on Visual Grounding}}

The results of ALBEF on VG task are shown in Table~\ref{tab_6}. We have the following main findings: (1) Consistent with the results on the previous tasks, perturbing bi-modal inputs is stronger than perturbing any single-modal input, and attacking full embedding outperforms attacking {\verb|[CLS]|} of embedding for perturbing image-modal input. (2) \texttt{Bi@Multi$\rm_{full}$} achieved the best performance in all settings of attacks. This further demonstrates the superiority of attacking multimodal embedding.


\begin{figure*}[t]
\begin{minipage}[b]{0.4\linewidth}
\centering
\includegraphics[width=0.99\textwidth]{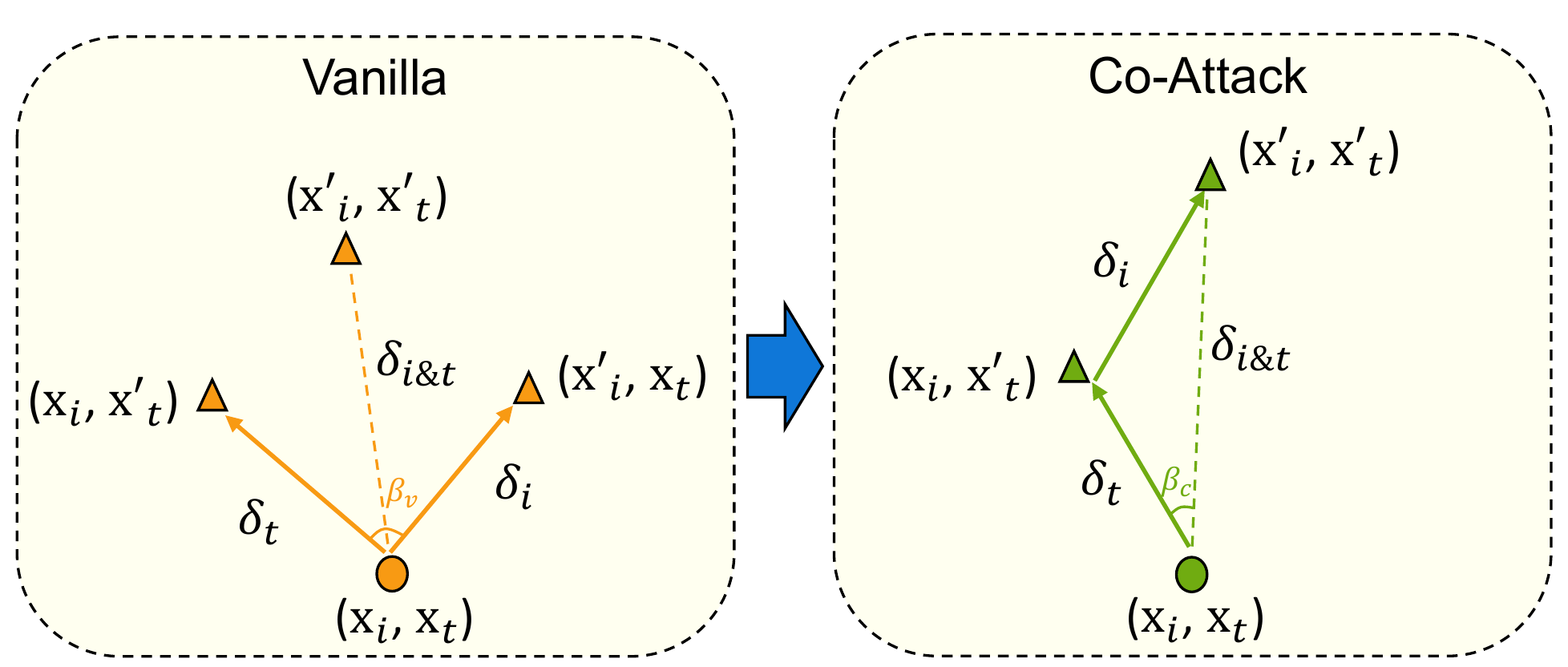}
\centerline{(a) multimodal embedding space (for fused VLP model)}
\end{minipage}
\begin{minipage}[b]{0.4\linewidth}
\centering
\includegraphics[width=0.99\textwidth]{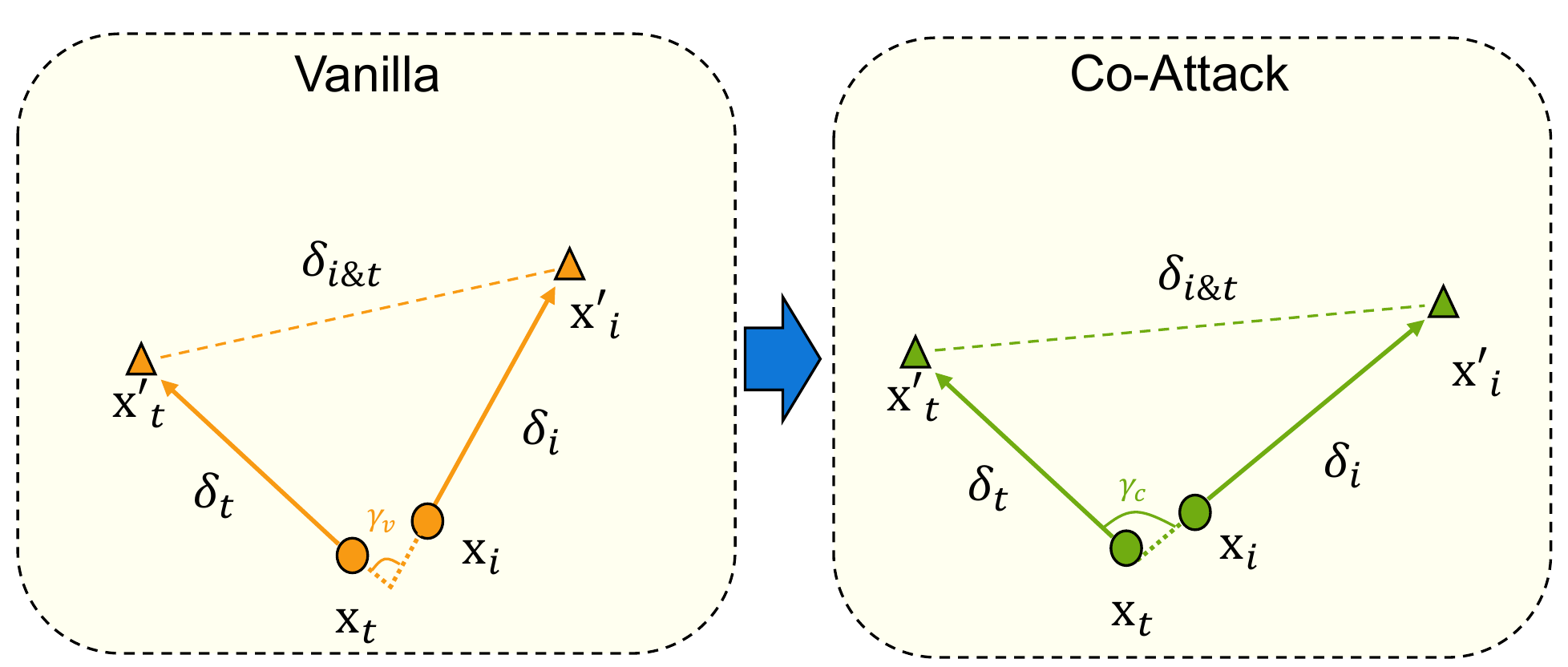}
\centerline{(b) unimodal embedding space (for aligned VLP model)}
\end{minipage}
\caption{Illustration of Co-Attack. $\delta_{i\&t}$ represents the resultant perturbation from image-modal perturbation $\delta_{i}$ and text-modal perturbation $\delta_{t}$ in embedding space. (a) In multimodal embedding space, $\beta$ represents the angle between $\delta_{i}$ and $\delta_{t}$. Co-Attack aims to reduce $\beta$ and enlarge $\delta_{i\&t}$. (b) In unimodal embedding space, $\gamma$ represents the angle between $\delta_{i}$ and $\delta_{t}$. Co-Attack aims to enlarge both $\gamma$ and $\delta_{i\&t}$. }\label{fig_3}
\end{figure*}

\begin{figure}[t]
\begin{minipage}[b]{0.49\linewidth}
\centering
\includegraphics[width=0.99\textwidth]{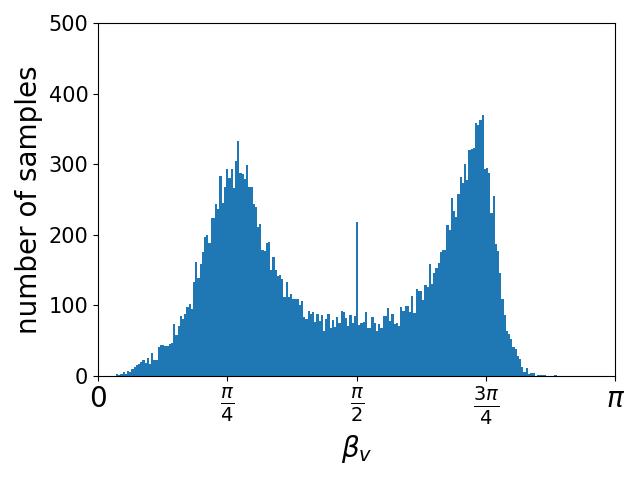}
\centerline{(a) Vanilla in multimodal}
\end{minipage}
\begin{minipage}[b]{0.49\linewidth}
\centering
\includegraphics[width=0.99\textwidth]{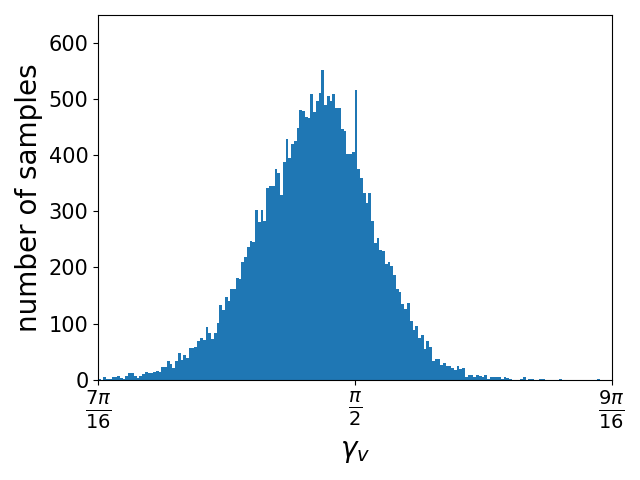}
\centerline{(b) Vanilla in unimodal}
\end{minipage}
\begin{minipage}[b]{0.49\linewidth}
\centering
\includegraphics[width=0.99\textwidth]{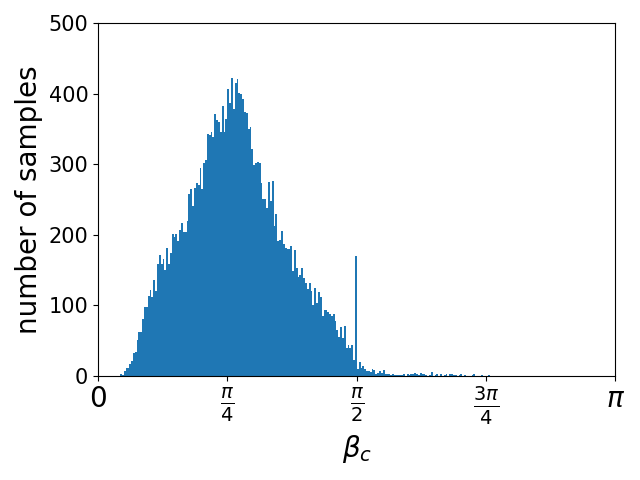}
\centerline{(c) Co-Attack in multimodal}
\end{minipage}
\begin{minipage}[b]{0.49\linewidth}
\centering
\includegraphics[width=0.99\textwidth]{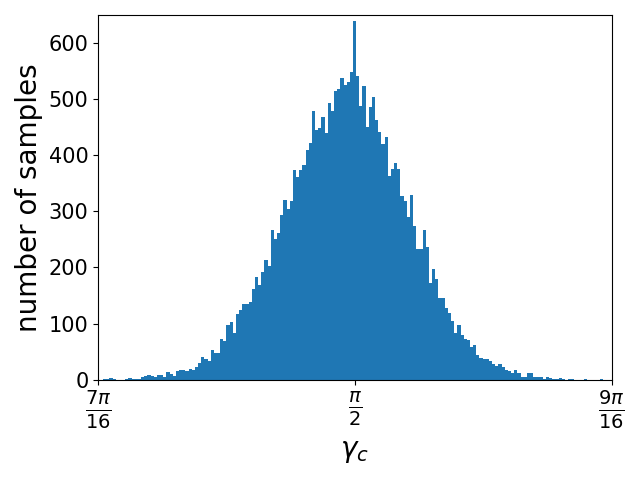}
\centerline{(d) Co-Attack in unimodal}
\end{minipage}

\caption{Statistical results on SNLI-VE of the angles between the image-modal attack and text-modal attack in the embedding space. (a) and (b) denote vanilla attack in multimodal embedding space and unimodal embedding space, respectively. (c) and (d) denote Co-Attack in multimodal embedding space and unimodal embedding space, respectively.}
 \label{fig_4}
\end{figure}


\paragraph{\textbf{Summary}}

We concluded some shared observations as guidance on designing multimodal adversarial attacks in VLP models. (1) Regarding the observations on different attack settings, perturbing bi-modal inputs is stronger than perturbing any single-modal input. This suggests that stronger adversarial attacks are expected if multiple modal inputs are allowed to be perturbed. In addition, for perturbing the single-modal input, {\verb|[CLS]|} of embedding has less impact on image modality compared to the text modality. For perturbing the bi-modal inputs, attacking {\verb|[CLS]|} of embedding is limited and even harmful compared to attacking the full embedding. (2) Regarding the observations on different downstream V+L tasks, for the tasks like image-text retrieval that require intermediate outputs (unimodal embedding) involved in the inference, the result of attacking the full embedding is consistent with attacking {\verb|[CLS]|} of embedding. But overall, attacking the multimodal embedding has a superior or comparable performance than the unimodal embedding along all tasks.

In addition, we derived some insights on constructing robust VLP models. Regarding the observations on different VLP models, the pre-training objectives of ALBEF (fused VLP model) and CLIP$_{\rm ViT}$ (aligned VLP model) contribute no apparently different impact on the adversarial robustness. For the models with alternative image encoders, such as CLIP, ViT is a superior choice compared to CNN for image encoder in terms of accuracy and robustness.


\section{Collaborative Multimodal Adversarial Attack in VLP Model}\label{sec_4}

\subsection{Methodology}

Although the above analysis found that perturbing the textual and visual modalities simultaneously is more effective than perturbing one modality singly. However, as discussed in Figure~\ref{fig_1}, there is a risk of leading to a countervailing $1+1<1$ effect from attacking both modalities independently. We address this issue by developing a collaborative multimodal adversarial attack solution called \emph{collaborative multimodal adversarial attack} (Co-Attack). This allows us to collectively carry out the attack on the image modality and the text modality. The aim of Co-Attack is to encourage the perturbed multimodal embedding away from the original multimodal embedding, or the perturbed image-modal embedding away from the perturbed text-modal embedding. Since Co-Attack can be adapted for attacking both multimodal and unimodal embeddings, it is applicable to both fused VLP and aligned VLP models.

\subsubsection{Attacking Multimodal Embedding}

For attacking multimodal embedding, Co-Attack attempts to collaboratively perturb the input text and input image, which encourages the perturbed multimodal embedding away from the original multimodal embedding. Figure~\ref{fig_3}(a) shows that without considering the consistency between the two attacks, the vanilla attack yields smaller resultant perturbation $\delta_{i\&t}$ and greater angle $\beta_{v}$ between text-modal perturbation $\delta_{t}$ and image-modal perturbation $\delta_{i}$. In contrast, Co-Attack collectively carries out the text-modal perturbation $\delta_{t}$ and image-modal perturbation $\delta_{i}$, achieving stronger resultant perturbation $\delta_{i\&t}$ and smaller angle $\beta_{c}$. As shown in Figure~\ref{fig_4}(a) and Figure~\ref{fig_4}(c), $\beta_{v}$ yielded by vanilla attack is distributed over $[0, \pi]$, and Co-Attack narrows the distribution and makes $\beta_{c}$ mainly distributions around $[0, \frac{\pi}{2}]$.

Next, we elaborate on how to realize Co-Attack. The main challenge of implementing a collaborative adversarial attack lies in the input representation gap between the continuous image modality and the discrete text modality. To solve this problem, we use a step-wise scheme that first perturbs the discrete inputs (text) and then perturbs the continuous inputs (image) given the text perturbation. The reason we first perturb text-modal input is that it is difficult to optimize the designed objective in discrete space. So we start with a text-modal perturbation and use it as a criterion, and then proceed to the image-modal perturbation. The adversarial text $\mathbf{x}^{'}_t$ can be derived by Equation~\eqref{eq4}. The adversarial attack on image modality is conducted by:
\begin{equation}\label{eq6}
\begin{aligned}
\max \ & \mathcal{L} (E_m(E_i(\mathbf{x}^{'}_i), E_t(\mathbf{x}^{'}_t)) , E_m(E_i(\mathbf{x}_i), E_t(\mathbf{x}^{'}_t))) \\ 
& + \alpha_1 \cdot \mathcal{L} (E_m(E_i(\mathbf{x}^{'}_i), E_t(\mathbf{x}^{'}_t)) , E_m(E_i(\mathbf{x}_i), E_t(\mathbf{x}_t))),
\end{aligned}
\end{equation}
where the second term corresponds to $\delta_{i\&t}$ in Figure~\ref{fig_3}(a), and $\alpha_1$ is a hyper-parameter that controls the contributions of the second term. The sensitivity to $\alpha_1$ is discussed in Section~\ref{sec4_2_3}. The above optimization problem can be easily solved by PGD-like procedures.

\subsubsection{Attacking Unimodal Embedding}

For attacking unimodal embedding, Co-Attack attempts to encourage the perturbed image-modal embedding away from the perturbed text-modal embedding. It should be noted that the unimodal embedding space is slightly different from the multimodal embedding space. As shown in Figure~\ref{fig_3}(b), the image-text sample pair corresponds to two different representations in the unimodal embedding space that are close together rather than one shared representation in the multimodal embedding space. It is shown that without considering the consistency between the two attacks, the vanilla attack yields both smaller resultant perturbation $\delta_{i\&t}$ and angle $\gamma_{v}$ between the text-modal perturbation $\delta_{t}$ and image-modal perturbation $\delta_{i}$. In contrast, Co-Attack collectively carries out the text-modal perturbation $\delta_{t}$ and image-modal perturbation $\delta_{i}$, achieving both greater resultant perturbation $\delta_{i\&t}$ and angle $\gamma_{c}$. As shown in Figure~\ref{fig_4}(b) and Figure~\ref{fig_4}(d), compared to the the $\gamma_{v}$ yielded by vanilla attack, Co-Attack shifts the $\gamma_{c}$ to a greater distribution.

Similar to attacking multimodal embedding, we first perturb input text to derive the adversarial text $\mathbf{x}^{'}_t$ according to Equation~\eqref{eq3}. Then similar to Equation~\eqref{eq6}, we perform the adversarial attack on image modality as follows:
\begin{equation}\label{eq7}
\max \ \mathcal{L} (E_i(\mathbf{x}^{'}_i) , E_i(\mathbf{x}_i))
 + \alpha_2 \cdot \mathcal{L} (E_i(\mathbf{x}^{'}_i) , E_t(\mathbf{x}^{'}_t)),
\end{equation}
where the second term corresponds to $\delta_{i\&t}$ in Figure~\ref{fig_3}(b), and $\alpha_2$ is a hyper-parameter that controls the contributions of the second term. The sensitivity to $\alpha_2$ is discussed in Section~\ref{sec4_2_3}.

\begin{table*}[t]
  \centering
  \caption{Comparison results on image-text retrieval. The reported value is attack success rate.}\label{tab_7}
  \begin{tabular}{cc|ccc|ccc|ccc|ccc}
  \toprule
  \multirow{3}{*}{Model} &\multirow{3}{*}{Attack} & \multicolumn{6}{c|}{Flickr30K (1K test set)} & \multicolumn{6}{c}{MSCOCO (5K test set)} \\
    & & \multicolumn{3}{c}{TR} & \multicolumn{3}{c|}{IR} & \multicolumn{3}{c}{TR} & \multicolumn{3}{c}{IR} \\
    \cline{3-14}
    & & R@1 & R@5 & R@10 & R@1 & R@5 & R@10 & R@1 & R@5 & R@10 & R@1 & R@5 & R@10 \\
    \hline
    \multirow{6}{*}{ALBEF}&Fooling VQA  & 12.80 & 4.70 & 3.10 & 12.72 & 6.62 & 4.40 & 39.68 & 31.50 & 24.82 & 34.23 & 34.73 & 29.70 \\
    &SSAP  & 66.10 & 55.30 & 49.60 & 63.54 & 64.82 & 60.62 & 54.30 & 58.58 & 55.32 & 46.77 & 60.26 & 60.94 \\
    &SSAP-MIM & 61.20 & 51.50 & 45.70 & 60.52 & 60.94 & 57.42 & 48.72 & 51.64 & 48.56 & 42.95 & 53.95 & 54.29 \\
    &SSAP-SI & 70.20 & 61.50 & 54.70 & 66.70 & 67.90 & 64.32 & 58.08 & 62.64 & 59.16 & 49.05 & 62.62 & 63.57 \\
    &Vanilla & 65.70 & 55.40 & 48.10 & 62.92 & 62.80 & 58.28 & 60.54 & 65.78 & 62.72 & 51.08 & 65.86 & 67.09 \\
    &Co-Attack & 70.60 & 60.50 & 53.50 & 67.22 & 67.10 & 62.76 & 58.84 & 64.18 & 61.40 & 51.13 & 66.17 & 67.73 \\
    & Co-Attack-SI & \textbf{72.20} & \textbf{63.50} & \textbf{58.20} & \textbf{69.72} & \textbf{71.00} & \textbf{67.12} & \textbf{65.90} & \textbf{71.74} & \textbf{67.92} & \textbf{54.39} & \textbf{70.16} & \textbf{71.59} \\
    \hline
    \multirow{6}{*}{CLIP$_{\rm ViT}$}& Fooling VQA & 9.90 & 5.60 & 3.40 & 8.26 & 5.54 & 4.20 & 14.52 & 13.68 & 12.10 & 6.58 & 8.55 & 8.02  \\
    & SSAP & 56.20 & 49.90 & 42.20 & 43.68 & 47.00 & 42.58 & 41.44 & 52.80 & 53.34 & 26.70 & 41.51 & 45.43 \\
    & SSAP-MIM & 48.10 & 42.20 & 35.30 & 38.40 & 39.42 & 35.52 & 38.40 & 48.10 & 47.70 & 24.37 & 37.47 & 40.74 \\
    & SSAP-SI & 58.90 & 53.90 & 46.70 & 46.72 & 53.54 & 51.20 & 43.44 & 57.06 & 58.50 & 27.22 & 43.38 & 48.33 \\
    & Vanilla & 64.00 & 60.90 & 52.90 & 51.08 & 60.34 & 58.54 & 46.34 & 62.10 & 64.34 & 29.75 & 48.78 & 54.88 \\
    & Co-Attack & 73.80 & \textbf{79.50} & 74.90 & \textbf{58.14} & \textbf{75.52} & \textbf{77.38} & 50.98 & \textbf{72.34} & \textbf{78.12} & \textbf{32.31} & \textbf{55.66} & \textbf{64.67} \\
    & Co-Attack-SI & \textbf{74.40} & 79.40 & \textbf{76.00} & 57.82 & 75.20 & 77.32 & \textbf{51.06} & 72.26 & 77.66 & 32.21 & 55.24 & 64.23 \\
  \bottomrule
\end{tabular}
\end{table*}

\begin{table*}[t]
  \centering
  \caption{Comparison results on visual entailment. The reported value is attack success rate.}\label{tab_8}
  \setlength{\tabcolsep}{3.8mm}{
  \begin{tabular}{c|ccccccc}
  \toprule
  \diagbox{Model}{Attack} & Fooling VQA & SSAP & SSAP-MIM & SSAP-SI & Vanilla & Co-Attack & Co-Attack-SI \\
  \hline
  ALBEF & 58.21 & 77.30 & 76.58 & 70.18 & 70.57 & \textbf{79.27} & 76.33 \\
  TCL & 51.88 & 72.47 & 71.80 & 68.89 & 66.46 & \textbf{76.43} & 74.94\\
  \bottomrule
\end{tabular}}
\end{table*}

\begin{figure}[t]
\begin{minipage}[b]{0.25\linewidth}
\centering
\includegraphics[width=0.99\textwidth]{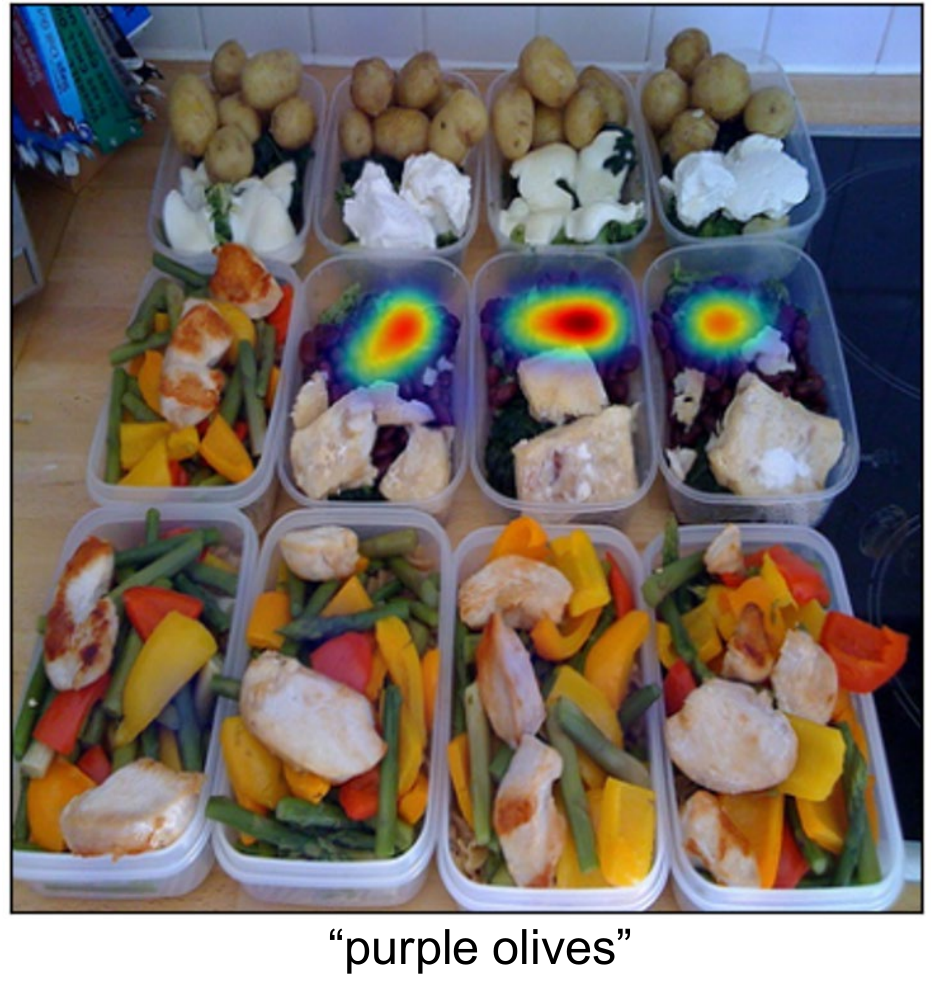}
\centerline{(a)original}
\end{minipage}
\begin{minipage}[b]{0.25\linewidth}
\centering
\includegraphics[width=0.99\textwidth]{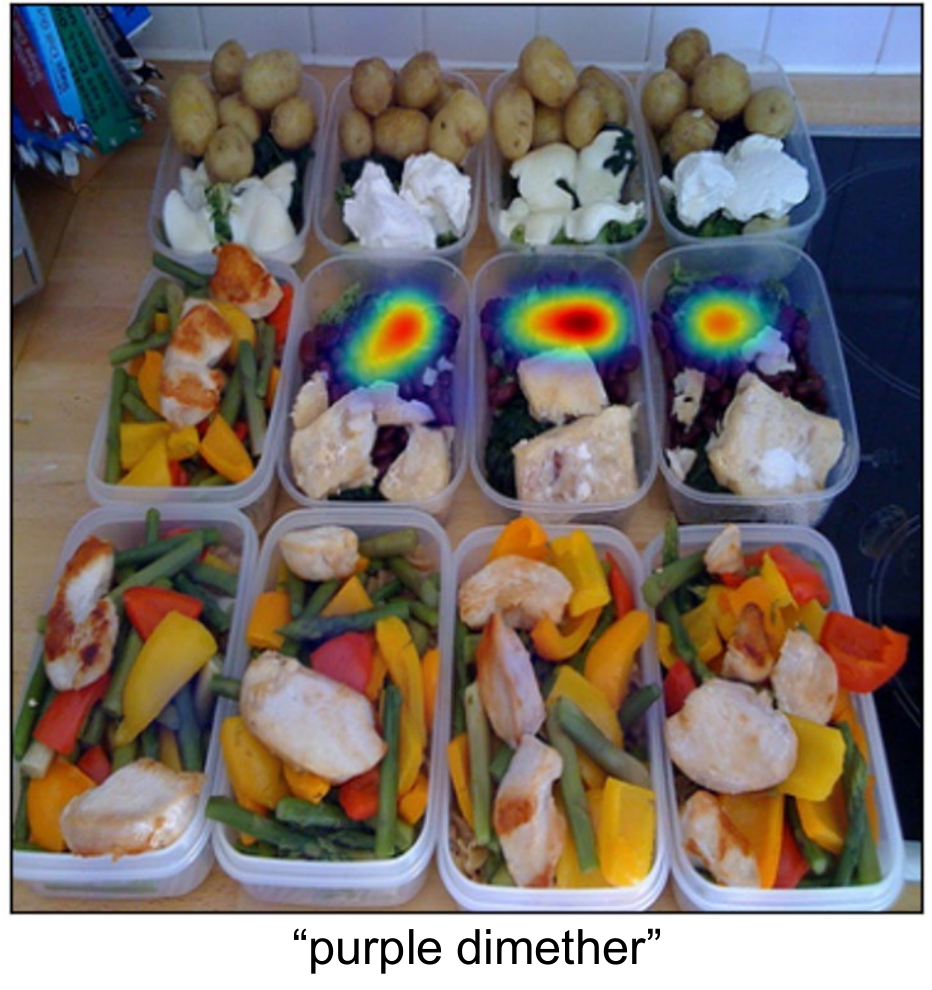}
\centerline{(b) text attack}
\end{minipage}
\begin{minipage}[b]{0.25\linewidth}
\centering
\includegraphics[width=0.99\textwidth]{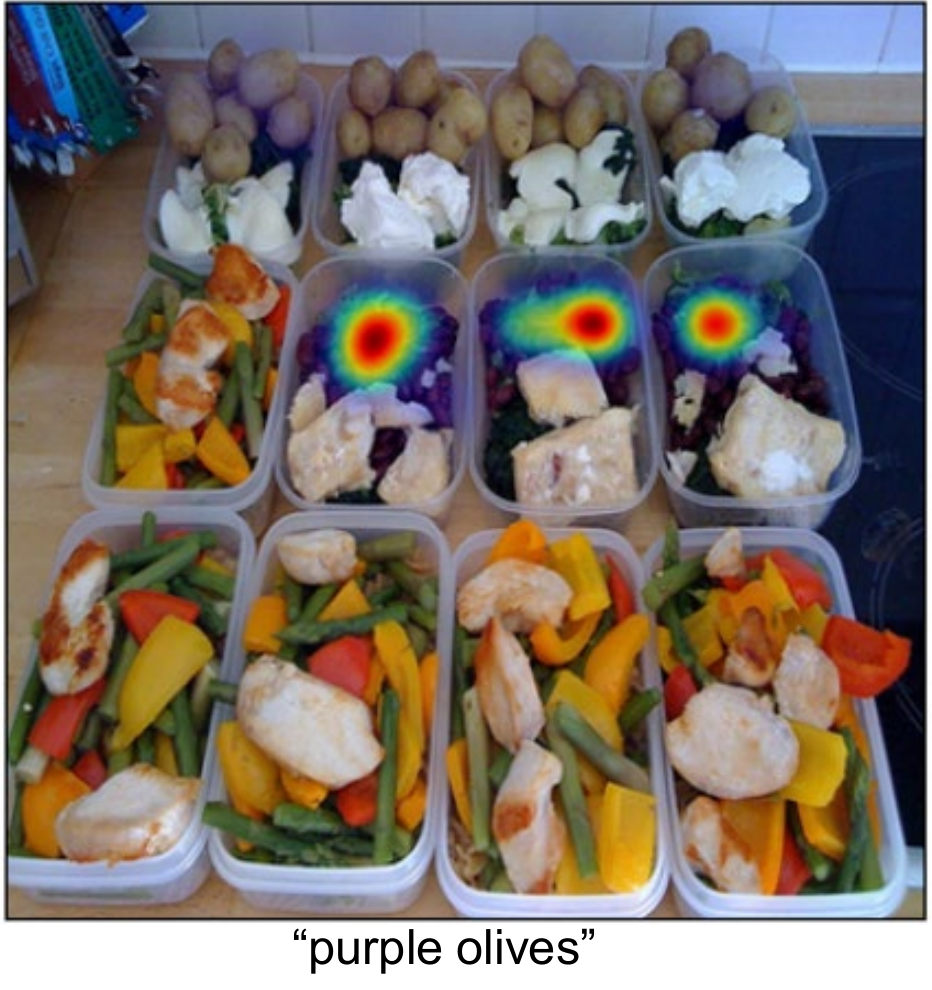}
\centerline{(c) image attack}
\end{minipage}
\begin{minipage}[b]{0.25\linewidth}
\centering
\includegraphics[width=0.99\textwidth]{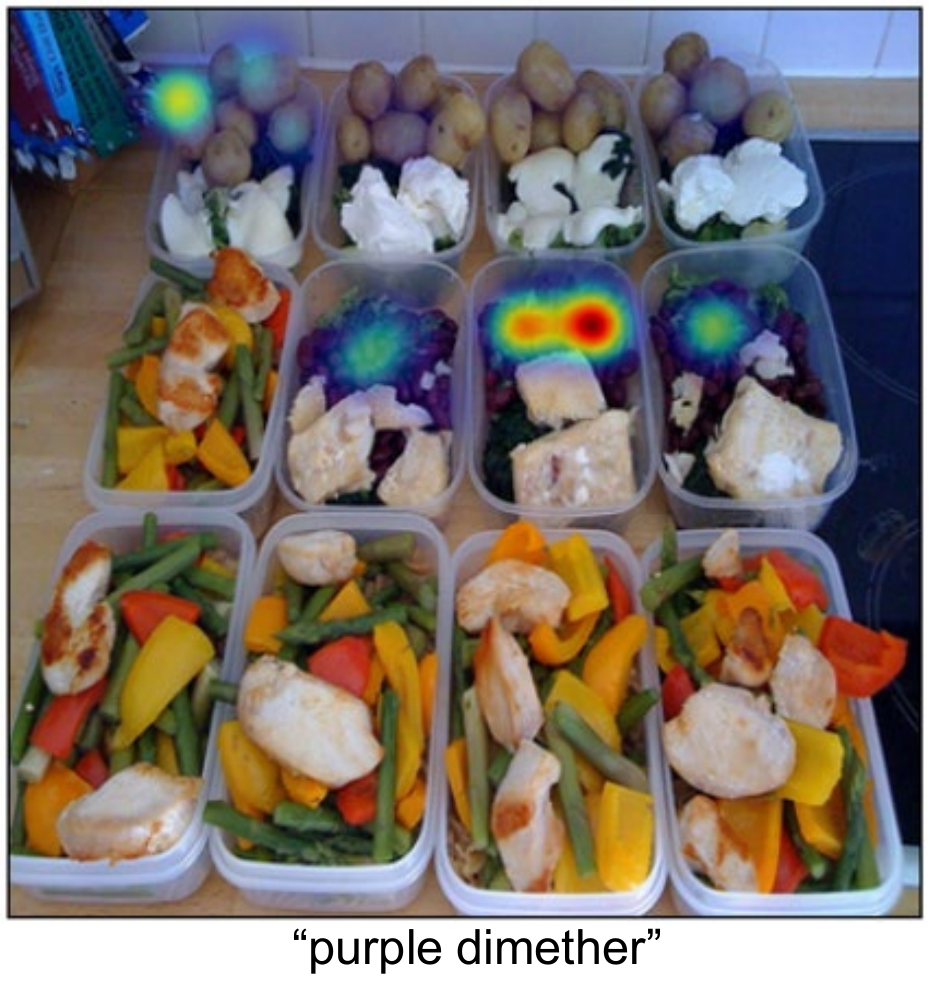}
\centerline{(d) vanilla attack}
\end{minipage}
\begin{minipage}[b]{0.25\linewidth}
\centering
\includegraphics[width=0.99\textwidth]{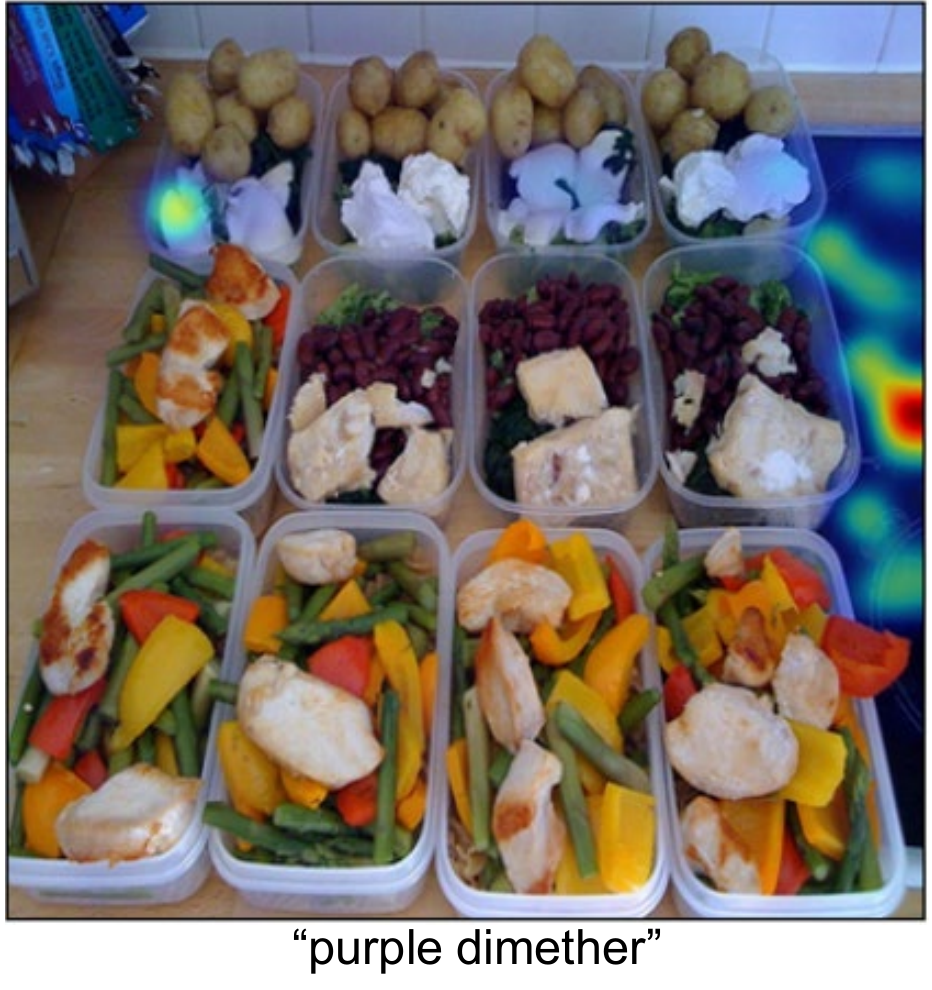}
\centerline{(e) Co-Attack}
\end{minipage}
\caption{The Grad-CAM visualizations of (a) the original sample pair ($\mathbf{x}_i, \mathbf{x}_t$), (b) ($\mathbf{x}_i, \mathbf{x}^{'}_t$) derived by \emph{Text@Multi$\rm_{full}$}, (c) ($\mathbf{x}^{'}_i, \mathbf{x}_t$) derived by \emph{Image@Multi$\rm_{full}$}, (d) ($\mathbf{x}^{'}_i, \mathbf{x}^{'}_t$) derived by \emph{Bi@Multi$\rm_{full}$} (vanilla attack), (e) ($\mathbf{x}^{'}_i, \mathbf{x}^{'}_t$) derived by Co-Attack.}\label{fig_5}
\end{figure}

\subsection{Experiments}\label{sec_42}

In this subsection, we used the same experimental settings as in Section~\ref{sec_31}. The momentum item for MIM is set to $0.9$, and the number of scale copies for SI is set to $5$. We set both $\alpha_1$ and $\alpha_2$ to $3$.

\subsubsection{Comparison Results}

To demonstrate the performance improvement of our methods, we compared the proposed Co-Attack with $5$ baseline attacks as follows.
\begin{itemize}
    \item \textbf{Fooling VQA} is an attack method for the classification problem (VQA model), which uses ADAM optimizer to solve the cross-entropy loss to add image noises~\cite{xu2018fooling}.
    \item \textbf{Single-Source Adversarial Perturbations (SSAP)} is used by \citet{yang2021defending} to evaluate adversarial robustness of VLP models, which is implemented by PGD to solve the cross-entropy loss to add image noises.
    \item \textbf{SSAP-MIM} and \textbf{SSAP-SI} are two baseline methods that replace PGD by introducing the more advanced MIM~\cite{dong2018boosting} and SI~\cite{lin2020nesterov} respectively, since the optimization algorithms in SSAP are replaceable. 
    \item \textbf{Vanilla} is the strongest attack analyzed in Section~\ref{sec_3}, which is also introduced as a baseline method, such as \texttt{Bi@Multi$\rm_{full}$} for the ALBEF model on the VE task.
\end{itemize}
Similarly, Co-Attack can also replace the PGD in the optimization algorithm with the SI, denoted as Co-Attack-SI. For a fair comparison, for the task that does not use cross-entropy, i.e., image-text retrieval, our method and all baseline methods attack embedding-wise representation as the target. For tasks that use cross-entropy, i.e. VE and VG, our method and all baseline methods attack logit-wise representation as the target.

Table~\ref{tab_7} shows the comparison results for the image-text retrieval task. Table~\ref{tab_8} presents the comparison results for the VE task. We can observe that: (1) Since Vanilla employs the strongest attack settings observed from our analysis in Section~\ref{sec_3}, Vanilla achieves basically superior attack performance over other baselines. This validates the reliability of the analysis results. (2) Co-Attack outperforms all baseline attacks. This demonstrates that Co-Attack improves the collaborative performance in attacking VLP models.

\subsubsection{Visualization Results}

To understand Co-Attack more intuitively, we provide the Grad-CAM visualizations for the VG task from ALBEF on RefCOCO+ dataset in Figure~\ref{fig_5}. Grad-CAM visualizations reflect the heat map where model would look when making decisions~\cite{selvaraju2017grad}. We noticed that perturbing single-modal input hardly changes the heat map. Vanilla slightly shifts the heat map, yet remaining noticeable regions on the object of interest. Co-Attack makes the model focus on the regions that deviate from the ground-truth so as to mislead the inference result.

\subsubsection{Ablation Study}\label{sec4_2_3}

\begin{figure}[t]
\begin{minipage}[b]{0.49\linewidth}
\centering
\includegraphics[width=0.99\textwidth]{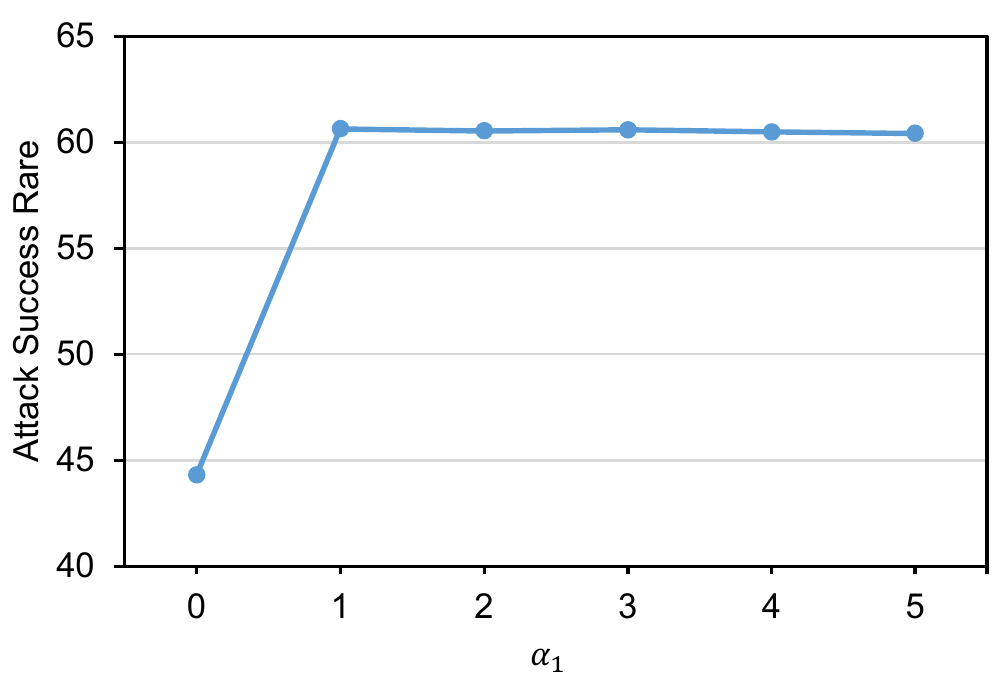}
\centerline{(a) The impact of $\alpha_1$}
\end{minipage}
\begin{minipage}[b]{0.49\linewidth}
\centering
\includegraphics[width=0.99\textwidth]{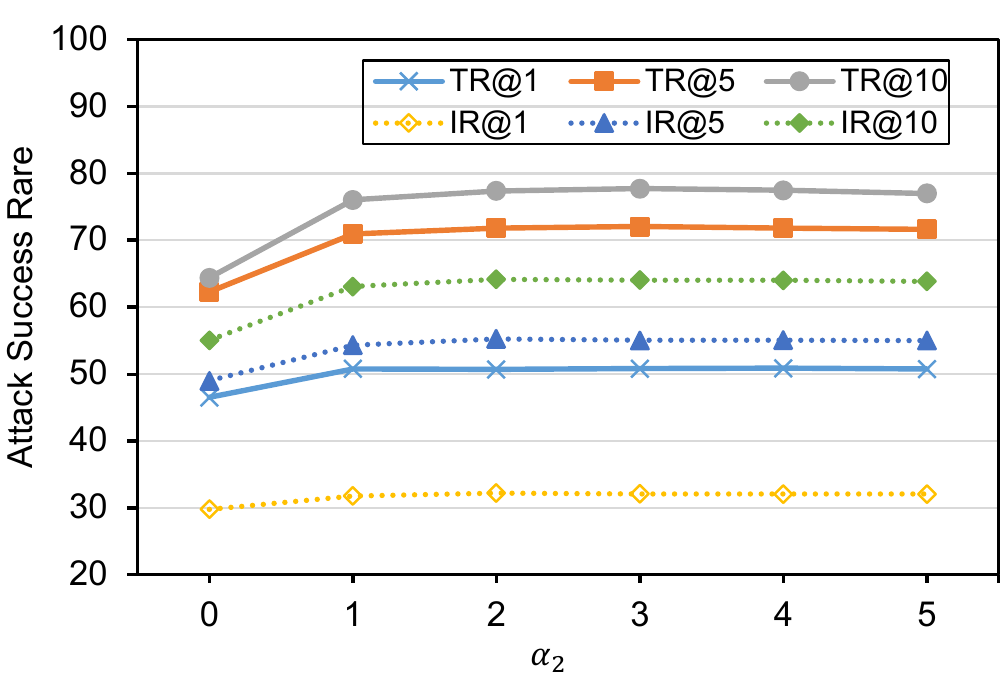}
\centerline{(b) The impact of $\alpha_2$}
\end{minipage}
\caption{The impact of $\alpha_1$ and $\alpha_2$. (a) The visual entailment result of ALBEF with different $\alpha_1$. (b) The image-text retrieval result of CLIP$\rm{_{ViT}}$ with different $\alpha_2$.}\label{fig_6}
\end{figure}

We conducted the ablation experiments to study the impact of $\alpha_1$ in Equation~\eqref{eq6} and $\alpha_2$ in Equation~\eqref{eq7}. We adjusted $\alpha_1$ and $\alpha_2$ within the range of $[0, 5]$ with a step of $1$ and examined their influences on ALBEF and CLIP$\rm{_{ViT}}$, respectively. The visual entailment result of ALBEF on SNLI-VE dataset is shown in Figure~\ref{fig_6}(a), and the image-text retrieval result of CLIP$\rm{_{ViT}}$ on COCO is shown in Figure~\ref{fig_6}(b). It is shown that as $\alpha_1 > 0$ and $\alpha_2 > 0$, the attack performance becomes stronger. This demonstrates the importance of the second term in Equation~\eqref{eq6} and Equation~\eqref{eq7}. The results are comparable when $\alpha_1 \geq 1$ and $\alpha_2 \geq 1$, indicating that Co-Attack is not sensitive to hyper-parameters and does not require elaborate tuning for the hyper-parameters.

\section{Conclusion}

In this paper, we investigated the adversarial attack on VLP models. First, we analyzed the performance of the adversarial attacks with different attack settings. Regarding the derived observations, we conclude the insights on designing the multimodal adversarial attacks and improving robustness in VLP models. Second, we developed a novel multimodal adversarial attack for VLP models. 

We hope this study can draw attention on the distinct properties of adversarial robustness of multimodal models.

\begin{acks}
This work is supported by the National Key R\&D Program of China (Grant No. 2018AAA0100604), the National Natural Science Foundation of China (Grant No. 61832002, 62172094), and Beijing Natural Science Foundation (No. JQ20023). The computing resources of Pengcheng Cloudbrain are used in this research.
\end{acks}

\bibliographystyle{ACM-Reference-Format}
\balance
\bibliography{sample}

\end{document}